\documentclass[11pt]{article}

\usepackage[final]{acl}

\usepackage{times}
\usepackage{latexsym}

\usepackage[T1]{fontenc}

\usepackage[utf8]{inputenc}

\usepackage{microtype}

\usepackage{inconsolata}

\usepackage{graphicx}
\usepackage{xcolor}

\usepackage{multirow}
\usepackage{placeins}
\usepackage{dblfloatfix}

\usepackage{amsmath}
\usepackage{booktabs}

\title{GenState-AI: State-Aware Dataset for Text-to-Video Retrieval on AI-Generated Videos}

\author{
\textbf{Minghan Li}$^{*,\dagger}$,
\textbf{Tongna Chen}$^{*}$,
\textbf{Tianrui Lv}$^{*}$\\
\textbf{Yishuai Zhang},
\textbf{Suchao An},
\textbf{Guodong Zhou}\\
School of Computer Science and Technology, Soochow University, Suzhou, China\\
$^{*}$The first three authors contributed equally. \quad
$^{\dagger}$Corresponding author.\\
\texttt{mhli@suda.edu.cn}
}

\begin{document}
\maketitle

\begin{abstract}
Existing text-to-video retrieval benchmarks often permit matching from static appearance, making it difficult to determine whether models capture state changes and final outcomes. We introduce GenState-AI\footnote{Dataset: \url{https://huggingface.co/datasets/Mh-Li/genstate-ai}}, a benchmark comprising 956 AI-generated videos and 195 verified state-change queries across Toy, Cartoon, Home, and Manipulation domains. Its construction uses prompt-controlled variants that target either conflicting temporal/state outcomes or semantic substitutions. Following human verification, all valid videos contribute to subset-level retrieval pools, while complete query-specific triplets support diagnostic hard-negative ordering. We evaluate VCF-Lik, Qwen3-VL, and LanguageBind using both standard Recall@K and triplet-based consistency metrics. Qwen3-VL (8B) achieves the best macro-average R@1 and triplet consistency, while LanguageBind obtains the highest semantic discrimination accuracy. However, stronger retrieval does not consistently yield correct hard-negative ordering, and Qwen3-VL and LanguageBind remain weak on Manipulation triplets, where VCF-Lik performs substantially better. GenState-AI therefore provides a controlled diagnostic testbed for distinguishing global retrieval quality from fine-grained temporal and semantic grounding in AI-generated videos.
\end{abstract}

\begin{figure*}[t]
  \centering
  \includegraphics[width=\textwidth]{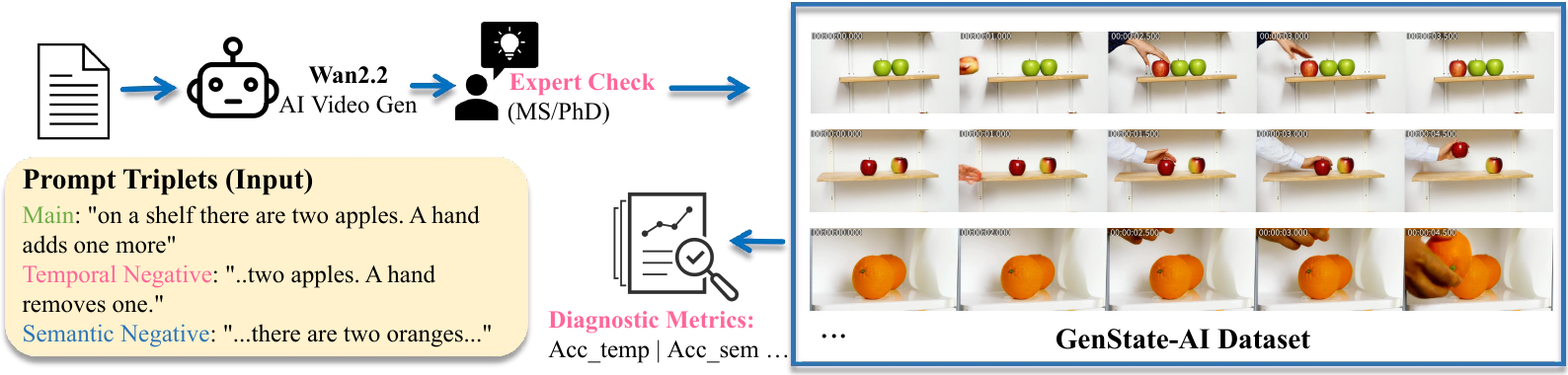}
  \caption{Overview of the GenState-AI construction and evaluation pipeline, including prompt-triplet construction, AI-based video generation, human verification and filtering, and retrieval evaluation.}
  \label{fig:pipeline}
\end{figure*}

\section{Introduction}
\label{sec:intro}

Text-to-video retrieval aims to find the most relevant video given a natural language query, or vice versa, from large-scale collections~\cite{miech2019howto100m, gabeur2020multi, luo2022clip4clip,ma2022x,reddy2025video,wang2024text,zhang2025text}. Recent work has shown that multi-modal large language models (MLLMs) can significantly improve retrieval, especially for long or compositional queries~\cite{linmm, chen2025vlm}. A recent highlight is BLiM~\cite{ko2025bidirectional}, which trains an MLLM to estimate bidirectional likelihoods (text given video and video features given text) and mitigates candidate prior bias using Candidate Prior Normalization (CPN), achieving strong performance on several popular real-world text--video benchmarks.

In parallel, modern video generative models such as Wan2.2-TI2V-5B~\cite{wan2025}, Step-Video-TI2V~\cite{huang2025step} and Make-A-Video~\cite{singer2022make}, among others~\cite{hacohen2024ltx,ma2024latte,zhou2022magicvideo,hong2022cogvideo,zhao2025controlvideo,liu2024sora}, can synthesize high-resolution, temporally coherent videos conditioned on text or image prompts.

These models make it possible to construct synthetic benchmarks that go beyond the biases and constraints of Internet videos, for example by enforcing controlled state transitions or non-physical but semantically precise scenarios.

However, most existing retrieval benchmarks, and thus most retrieval methods implicitly assume that
(1) a single frame conveys most of the semantics, and
(2) videos follow natural visual statistics.
Consequently, negative samples in real-world datasets often differ from positives in many uncontrolled ways, allowing models to succeed through appearance matching rather than genuine temporal reasoning.
Under these conditions, models need not explicitly track how visual states change {over time}.

This limitation raises a fundamental question:
\begin{quote}
{How can we reliably evaluate whether retrieval models are genuinely state- and temporal-aware, especially in the context of controllable AI-generated videos?}
\end{quote}

To address this gap, we construct prompt-controlled hard negatives that target specific state or semantic conflicts, while acknowledging that independently generated videos may retain low-level visual variations.

\paragraph{GenState-AI}
We introduce {GenState-AI}, an AI-generated text-to-video retrieval benchmark explicitly designed around controlled state transitions (the process is shown in Fig.~\ref{fig:pipeline}).
We leverage the open-source Wan2.2-TI2V-5B hybrid text-image-to-video model~\cite{wan2025}, which can generate 720p videos at 24 fps,
to procedurally create a large collection of short clips.
Each clip is defined by a simple, human-readable prompt describing:
(1) a small set of tabletop objects,
(2) an initial configuration, and
(3) a final configuration after a short motion or state change.

Crucially, we also construct {hard negative} prompts through minimal edits that keep the intended scene, objects, and configuration fixed at the prompt level while targeting a conflicting {final state}.
For example, the teddy bear may move slightly but ultimately fails to cross to the other side of the mug.
Such pairs create retrieval scenarios where neither global appearance nor superficial motion cues are sufficient; models must accurately infer the {resulting state} of the scene to succeed.
Because the clips are independently generated, low-level visual appearance is not perfectly controlled; we therefore rely on human verification and diagnostic triplet analyses to reduce confounding effects.

We conduct a systematic evaluation of MLLM-based text-to-video retrieval on the proposed GenState-AI benchmark using four retrieval systems from three representative model families: (i) a likelihood-scoring T2V baseline built upon \textbf{VideoChat-Flash-Qwen2-7B\_res448}, (ii) \textbf{Qwen3-VL} scale-matched embedding--reranking baselines with 2B and 8B embedding/reranker configurations, and (iii) a \textbf{LanguageBind} video--text embedding baseline.
Despite their architectural differences, these baselines exhibit consistent failure patterns in synthetic, state-centric settings.
When evaluated on prompt-controlled AI-generated videos, they often rank candidates with globally ``on-topic'' appearance or temporally plausible but end-state-incorrect trajectories ahead of the true match, indicating insufficient grounding to decisive end-state evidence.
Scaling the Qwen3-VL configuration from 2B to 8B, with scale-matched embedding and reranking models, improves the macro-average R@1 from 0.539 to 0.581, while Acc$_{\text{tri}}$ increases only modestly from 0.693 to 0.721, showing that GenState-AI is sensitive to model capability while still exposing residual temporal-ordering failures.
However, this improvement is not uniform: Qwen3-VL (8B) still struggles on \textit{Manipulation} (R@1: 0.304; Acc$_{\text{tri}}$: 0.524), while VCF-Lik achieves much stronger triplet consistency on the same subset (0.810).

The triplet design exposes a consistent asymmetry across baselines: temporal hard negatives are harder to reject than semantic substitutions, even when the correct video ranks near the top.
This gap suggests that strong Recall@K does not necessarily imply reliable state-transition grounding, as models may favor globally ``on-topic'' or temporally plausible clips over evidence of the correct end state.

We further provide subset-level and triplet-order analyses to characterize temporal versus semantic confusions, highlighting open challenges for state- and time-sensitive text-to-video retrieval.

\section{Related Work}

Text-to-video retrieval has traditionally relied on dual encoders that align video and text representations through contrastive learning~\cite{sun2019videobert,miech2019howto100m,gabeur2020multi,luo2022clip4clip}.
More recent systems incorporate stronger multimodal encoders, rerankers, or generative likelihood scoring~\cite{hur2025narrating,chen2023videollm,girdhar2023imagebind,li2024videochat,li2026qwen3,zhu2024languagebind}.
For long-document text retrieval~\cite{li2025survey}, prior work has explored BERT-based dense intra-ranking and contextualized late interaction~\cite{li2022bert}, as well as key-block selection with local pre-ranking~\cite{li2023power}. More recent LLM-based rerankers select and compact query-relevant evidence before scoring~\cite{li2024evirerank}.
These model-centric advances improve retrieval on common datasets, but standard benchmarks such as MSR-VTT, DiDeMo, ActivityNet-Captions, and VATEX~\cite{xu2016msr,anne2017localizing,krishna2017dense,wang2019vatex} are dominated by real-world footage where global appearance often suffices.
In text-only dense retrieval, cross-domain generalization has also been studied through self-supervised pseudo-relevance labeling~\cite{li2024domain}.

Recent diagnostic benchmarks probe temporal and fine-grained video understanding more directly. RTime~\cite{du2024reversed} constructs time-reversed hard negatives, CaReBench~\cite{xu2024carebench} separates spatial and temporal annotations, and TempCompass~\cite{liu2024tempcompass} evaluates event order and attribute changes.
Synthetic and generated-video datasets have also emerged~\cite{liu2023genphys,chen2022learning,guo2024sparsectrl,wan2025}, but they primarily target physics, controllability, or generation quality rather than retrieval.
GenState-AI is complementary: it uses AI-generated clips and paired temporal/semantic hard negatives to diagnose whether retrieval models can distinguish the intended state outcome from visually plausible alternatives.

\section{GenState-AI Dataset}

\subsection{Design Principles}

GenState-AI is built around three principles:

\begin{enumerate}
\item \textbf{State-centric semantics.}
Each video is defined by a clear and observable change in state. One category of state transitions arises from the actor manipulating objects, such as moving a target object to alter its position or modifying the quantity of objects by adding or removing instances. Another category is driven by the actor’s own positional change, where the actor’s movement within the scene constitutes an independent form of state transition. Together, these two types of state changes enable the dataset to systematically capture action-induced state transformations.

\item \textbf{Minimalist prompts.}
We employ concise and unambiguous prompts that focus solely on the target state transition. Avoiding long or stylistically complex descriptions prevents semantic drift and ensures that both the model and the evaluation concentrate on the intended change of state.

\item \textbf{Hard negatives by state misalignment.}

For each main query, we construct hard-negative prompts through minimal prompt edits that target either a conflicting end-state or a semantic substitution while preserving the intended scene and object configuration at the prompt level.
These hard negatives fall into two categories. 
First, \textbf{semantic hard negatives} modify the target entity: for example, in tasks where an actor is supposed to move a specific object, the hard negative instead moves a different object; in scenarios where the actor's own movement is defined relative to a reference object, we replace the reference to induce a semantic mismatch. 
Second, \textbf{temporal hard negatives} alter the action dynamics themselves, such as keeping the object stationary when the positive sample requires movement, or performing only a partial motion that fails to reach the final state. 
Because the clips are generated independently from text-only prompts, low-level visual appearance is not perfectly controlled: background layout, lighting, object texture, pose, and generation artifacts may vary across videos in a triplet.
We therefore rely on human verification and triplet-level diagnostics to reduce confounding effects and to ensure that each retained hard negative remains an interpretable contrast to the main query.
These prompt-controlled hard negatives encourage models to rely on state reasoning rather than superficial topic cues, enabling a more stringent and fine-grained assessment of state understanding.

\end{enumerate}

\subsection{Video Generation with Wan2.2-TI2V-5B}

\begin{table}[t]
\centering
\small
\setlength{\tabcolsep}{6pt}
\caption{
Number of query descriptions per video category in our GenState-AI dataset, along with the average token length, computed using the Qwen2.5-1.5B tokenizer.}
\label{tab:text_counts}
\begin{tabular}{lcc}
\toprule
\textbf{Category} & \textbf{Number of Queries} & \textbf{Avg. Tokens} \\
\midrule
Home         & 127 & 20.1 \\
Toy          & 21  & 26.8 \\
Cartoon      & 24  & 30.7 \\
Manipulation & 23  & 14.7 \\
\midrule
\textbf{Total} & \textbf{195} & \textbf{21.5} \\
\bottomrule
\end{tabular}
\end{table}

We use the Wan2.2-TI2V-5B model~\cite{wan2025}, an open-source text-image-to-video generator that supports 1280$\times$704 resolution at 24 fps with a 5B-parameter hybrid architecture.

Our video dataset contains four categories: {Toy}, {Cartoon}, {Home}, and {Manipulation}, with a total of 956 videos. 
Each video has a resolution of $1280\times704$ (720p) at 24\,FPS and lasts approximately 5 seconds (121 frames). The {Cartoon} and {Toy} categories are non-realistic, synthetic scenarios designed to test models on stylized and imaginative content, 
while the {Home} and {Manipulation} categories include realistic scenes covering everyday objects and natural environments. 
By combining both realistic and non-realistic videos, our dataset provides a diverse benchmark for evaluating text-to-video retrieval models across various visual domains. 
Table~\ref{tab:text_counts} summarizes the number of main query descriptions for each video category in our GenState-AI dataset (after manual verification and filtering). 
The dataset contains four categories ({Home}, {Toy}, {Cartoon} and {Manipulation}), 
with a total of 195 main query descriptions. Notably, the {Home} category has the largest number of main queries, while {Toy} and {Cartoon}/{Manipulation} have fewer.

In our default setting, we generate 5-second clips at 720p using text-only conditioning, focusing on stable scenes (e.g., home, desk, kitchen counter).
Since all clips are independently generated, GenState-AI should be understood as a prompt-controlled benchmark rather than a pixel-controlled counterfactual video benchmark.

GPT-4 generated the prompt triplets under a shared standardized instruction.

Prompts follow a simple, controlled structure designed to isolate semantic and temporal variations:
\begin{itemize}
    \item {Main video example:} ``On a shelf there are two apples. A hand adds one more.''
    \item {Semantic hard negative:} ``On a shelf there are two oranges. A hand adds one more.''
    \item {Temporal hard negative:} ``On a shelf there are two apples. A hand removes one.''
\end{itemize}

{This prompt structure is designed to encourage coherent motion while avoiding complex interactions such as pouring, cutting, or complex human actions. Clips inconsistent with the intended prompt or target state are removed through manual verification.}

Table~\ref{tab:dataset_stats} summarizes the statistics of each category.

\begin{table}[t]
\centering
\caption{
Statistics of our video dataset. 
Each video has a resolution of $1280\times704$ (720p) at 24\,FPS and lasts about 5\,seconds (121 frames).
}
\label{tab:dataset_stats}
\resizebox{0.98\columnwidth}{!}{%
\begin{tabular}{lccccc}
\toprule
\textbf{Category} & \textbf{\#Videos} & \textbf{Resolution} & \textbf{FPS} & \textbf{Duration (s)} & \textbf{Size} \\
\midrule
Home  & 368 & $1280\!\times\!704$ & 24 & 5 & 2.53 GB \\ 
Toy        & 221 & $1280\!\times\!704$ & 24 & 5 & 1.01 GB \\
Cartoon    & 150 & $1280\!\times\!704$ & 24 & 5 & 522 MB \\
Manipulation     & 217 & $1280\!\times\!704$ & 24 & 5 & 1.92 GB \\
\midrule
\textbf{Total} & \textbf{956} &  &  & & \textbf{5.98 GB} \\
\bottomrule
\end{tabular}%
}
\end{table}

\subsection{Example Prompts with AI-Generated Videos}

Appendix~\ref{sec:appendix_examples} presents representative examples from each group of our benchmark.
For every query prompt, the examples show the corresponding main video alongside its temporal and semantic hard negatives.
These triplets illustrate the core difficulty of GenState-AI: despite being visually similar, the negatives differ from the main video either in {object identity} or in {state evolution}, requiring retrieval models to possess both semantic understanding and fine-grained temporal reasoning.

\subsection{Manual Check and Correction}
\label{sec:manual_check}

Although Wan2.2-TI2V-5B provides controllable video generation, it can occasionally produce clips that deviate from the intended prompt or fail to realize the target state transition.
To ensure the reliability of GenState-AI, we conduct a structured human verification process.
Five annotators (four M.S.\ students in AI and one Ph.D.\ researcher) divided the clips into five non-overlapping subsets, with exactly one reviewer per clip. They applied shared criteria for prompt consistency, required objects and actions, visibility of the intended transition, and the absence of severe visual artifacts; this protocol does not support inter-annotator agreement.

\textbf{Main-clip verification (query validity).}
For each query description, annotators first inspect its \textit{main} clip and verify whether the video faithfully expresses the intended state-change description.
If the main clip is incorrect or ambiguous, we discard \emph{only} this main clip (and regenerate a new main candidate if needed), because an invalid main clip would invalidate the query--video alignment and make subsequent triplet-based evaluation uninterpretable.

\textbf{Hard negatives as global distractors.}
In contrast, we do not enforce a strict \emph{pairwise} constraint between a hard negative and a particular main clip during manual checking.
Temporal and semantic hard-negative clips are retained even when the corresponding main clip is discarded, since these clips remain valid videos that can serve as \emph{global negatives} (distractors) for other queries in the candidate pool.
This design choice both increases the diversity of negatives and explains why the total number of verified clips is not necessarily a multiple of three.

\textbf{Triplet integrity without asymmetric replacement.}
{When forming query-specific triplets for triplet-ordering analyses, we retain only query groups for which the main, temporal, and semantic clips are all present in the evaluation export. We do not replace a missing component with a clip from another query, because that would introduce asymmetric examples and could bias retrieval evaluation. The resulting export contains 154 complete triplets; the remaining 41 main-query descriptions remain part of the full candidate-pool retrieval evaluation but are excluded from triplet-level metrics.}

Across approximately 1{,}800 generated video clips, a total of 956 valid clips remain after manual verification.
The total time to generate a single video on an RTX 5090 is approximately 7 minutes.
These verified clips support both (i) constructing a core set of query-specific triplets for triplet-based evaluation and (ii) enriching the candidate pool with additional global negative videos for more challenging retrieval.

\begin{table*}[!t]
\centering
\scriptsize
\setlength{\tabcolsep}{3pt}
\renewcommand{\arraystretch}{0.86}
\caption{Performance comparison of the baselines across different subsets on GenState-AI. The Avg.\ row reports the macro average over the four subsets. Best results are highlighted in \textbf{bold}. {Recall@K uses all videos in each corresponding subset pool and retains all 195 queries; the three triplet accuracies use the 154 complete triplets and exclude 41 incomplete groups.}}
\resizebox{0.80\textwidth}{!}{
\begin{tabular}{llcccccc}
\toprule
Subset & Baseline & R@1 & R@5 & R@10 & Acc$_{\text{temp}}$ & Acc$_{\text{sem}}$ & Acc$_{\text{tri}}$ \\
\midrule
\multirow{4}{*}{Toy}
 & VCF-Lik & 0.381 & 0.714 & 0.857 & {0.619} & {0.810} & {0.524} \\
 & Qwen3-VL (2B) & {0.571} & {\textbf{0.952}} & {0.952} & {0.667} & {\textbf{0.905}} & {0.667} \\
 & Qwen3-VL (8B) & {\textbf{0.667}} & {\textbf{0.952}} & {\textbf{1.000}} & {\textbf{0.762}} & {\textbf{0.905}} & {\textbf{0.714}} \\
 & LanguageBind & 0.571 & 0.905 & 0.952 & 0.619 & 0.857 & 0.619 \\
\midrule
\multirow{4}{*}{Cartoon}
 & VCF-Lik & {0.250} & {0.708} & {0.708} & {0.500} & {0.833} & {0.417} \\
 & Qwen3-VL (2B) & {0.708} & {\textbf{1.000}} & {\textbf{1.000}} & {0.792} & {0.875} & {0.750} \\
 & Qwen3-VL (8B) & {0.667} & {\textbf{1.000}} & {\textbf{1.000}} & {0.792} & {0.958} & {0.750} \\
 & LanguageBind & \textbf{0.833} & 0.958 & \textbf{1.000} & \textbf{0.833} & \textbf{1.000} & \textbf{0.833} \\
\midrule
\multirow{4}{*}{Home}
 & VCF-Lik & 0.386 & 0.512 & 0.591 & {0.682} & {0.920} & {0.648} \\
 & Qwen3-VL (2B) & {0.528} & {0.850} & {0.953} & {0.852} & {0.943} & {0.830} \\
 & Qwen3-VL (8B) & {\textbf{0.685}} & {\textbf{0.929}} & {\textbf{0.984}} & {\textbf{0.909}} & {0.977} & {\textbf{0.898}} \\
 & LanguageBind & 0.441 & 0.890 & 0.969 & 0.716 & \textbf{0.989} & 0.716 \\
\midrule
\multirow{4}{*}{Manipulation}
 & VCF-Lik & \textbf{0.478} & 0.870 & 0.957 & {\textbf{0.952}} & {0.810} & {\textbf{0.810}} \\
 & Qwen3-VL (2B) & {0.348} & {0.783} & {\textbf{0.957}} & {0.524} & {0.762} & {0.524} \\
 & Qwen3-VL (8B) & {0.304} & {\textbf{0.957}} & {\textbf{0.957}} & {0.619} & {\textbf{0.857}} & {0.524} \\
 & LanguageBind & 0.391 & 0.913 & 0.913 & 0.571 & \textbf{0.857} & 0.524 \\
\midrule
\multirow{4}{*}{Avg.}
 & VCF-Lik & {0.374} & {0.701} & {0.778} & {0.688} & {0.843} & {0.599} \\
 & Qwen3-VL (2B) & {0.539} & {0.896} & {0.965} & {0.709} & {0.871} & {0.693} \\
 & Qwen3-VL (8B) & {\textbf{0.581}} & {\textbf{0.960}} & {\textbf{0.985}} & {\textbf{0.770}} & {0.924} & {\textbf{0.721}} \\
 & LanguageBind & 0.559 & 0.916 & 0.958 & 0.685 & \textbf{0.926} & 0.673 \\
\bottomrule
\end{tabular}
}
\label{tab:genstate_triplet_metrics}
\end{table*}

\section{Evaluation Protocol}

\subsection{Standard Retrieval Setup}

{For each text query $q$, its mapped main clip $v^{\text{main}}$ is the sole relevant target. Recall@K ranks it against all videos in the corresponding evaluation pool; other main clips and all available temporal or semantic hard negatives are ordinary non-relevant candidates.}

We adopt standard text-to-video retrieval metrics.
To emphasize state-awareness, we also report the proposed triplet metrics and analyze error types.

\subsection{MLLM Baselines on GenState-AI}

We evaluate representative MLLM-based retrieval baselines on GenState-AI from three complementary paradigms.
First, we build a likelihood-scoring text-to-video baseline on top of \textbf{VideoChat-Flash-Qwen2-7B\_res448} (denoted as \textbf{VCF-Lik}).
For each candidate video $v$, we extract frame-level visual tokens with a frozen vision backbone and condition the MLLM on these tokens to score a text query $q$ via teacher forcing.
Concretely, we use the length-normalized token log-likelihood (equivalently, the negative cross-entropy) as the relevance score,
\begin{equation}
s_{\text{lik}}(q,v) = \frac{1}{|q|}\sum_{i=1}^{|q|} \log p_\theta(q_i \mid q_{<i}, v),
\end{equation}
and rank candidates by $s_{\text{lik}}(q,v)$.

Second, we adopt a two-stage pipeline based on \textbf{Qwen3-VL}, where \textbf{Qwen3-VL-Embedding} performs first-stage retrieval and \textbf{Qwen3-VL-Reranker} conducts second-stage re-ranking~\cite{li2026qwen3}.
{We evaluate scale-matched 2B and 8B embedding/reranker configurations; we denote them as \textbf{Qwen3-VL (2B)} and \textbf{Qwen3-VL (8B)}.}

Finally, we include \textbf{LanguageBind}~\cite{zhu2024languagebind}, a contrastive multimodal representation model that aligns video and text in a shared embedding space through language-based semantic alignment.
We use the fine-tuned video checkpoint \textit{LanguageBind\_Video\_FT} and rank candidates by text--video embedding similarity.
This baseline represents a single-stage dual-encoder retrieval paradigm without an explicit cross-candidate reranking step.

\subsection{Triplet Hard-Negative Ordering Metrics}
\label{sec:tripletMetrics}
{For a query whose mapped main, temporal-negative, and semantic-negative clips are all available, GenState-AI defines the query-specific \emph{triplet}}
\[
\bigl(v^{\text{main}},\;v^{\text{temp}},\;v^{\text{sem}}\bigr),
\]
where $v^{\text{main}}$ is the ground-truth video, $v^{\text{temp}}$ is a temporal hard negative, and $v^{\text{sem}}$ is a semantic hard negative.
{Given any retrieval model that produces a scalar score $s(q,v)$ (e.g., a likelihood, embedding-similarity, or reranking score), we can study not only the global rank of $v^{\text{main}}$, but also its \emph{relative ordering} with respect to the two hard negatives.}

We distinguish two evaluation contexts.
{Recall@K is computed from the standard retrieval ranking over all videos in the corresponding evaluated subset.}
In contrast, the following triplet metrics are computed as a diagnostic ordering task over only the query-specific triplet
\[
\mathcal{T}_i=\{v_i^{\text{main}},v_i^{\text{temp}},v_i^{\text{sem}}\}.
\]
That is, each model scores the three paired clips for the same query and the three clips are ranked relative to one another.
{For two-stage systems, this diagnostic evaluation bypasses the first-stage $K=20$ candidate filter and directly scores the three triplet clips with the final scoring or reranking module.}
Therefore, the triplet metrics are not mathematically constrained by full-pool Recall@K.

{Recall@K retains all 195 query descriptions. The triplet accuracies and relative-order histograms use only the 154 queries with complete main, temporal-negative, and semantic-negative mappings; 41 incomplete groups are excluded.}

Let $\mathrm{rank}_{\mathcal{T}}(q_i,v)$ denote the rank (1 is best) of candidate $v$ among the three videos in $\mathcal{T}_i$.
We define the following GenState-AI specific metrics:

\paragraph{Main-vs-Temporal accuracy.}
The probability that the model correctly prefers the main video over its temporal hard negative:
\begin{equation}
\begin{aligned}
\mathrm{Acc}_{\mathrm{temp}}
&= \frac{1}{N} \sum_{i=1}^{N}
\mathbf{1}\Bigl[
\mathrm{rank}_{\mathcal{T}}(q_i, v_i^{\mathrm{main}}) \\
&\qquad\qquad <
\mathrm{rank}_{\mathcal{T}}(q_i, v_i^{\mathrm{temp}})
\Bigr].
\end{aligned}
\end{equation}
{In the reported triplet diagnostics, $N=154$, corresponding to the queries with complete main, temporal-negative, and semantic-negative mappings.}

\paragraph{Main-vs-Semantic accuracy.}
Analogously, the probability that the main video is ranked ahead of its semantic hard negative:
\begin{equation}
\begin{aligned}
\mathrm{Acc}_{\mathrm{sem}}
&= \frac{1}{N} \sum_{i=1}^{N}
\mathbf{1}\Bigl[
\mathrm{rank}_{\mathcal{T}}(q_i, v_i^{\mathrm{main}}) \\
&\qquad\qquad <
\mathrm{rank}_{\mathcal{T}}(q_i, v_i^{\mathrm{sem}})
\Bigr].
\end{aligned}
\end{equation}
\paragraph{Triplet consistency.}
A stricter metric that requires the main video to be ranked ahead of \emph{both}
hard negatives. For brevity, let
$r_i^{\mathrm{m}} = \operatorname{rank}_{\mathcal{T}}(q_i, v_i^{\text{main}})$,
$r_i^{\mathrm{t}} = \operatorname{rank}_{\mathcal{T}}(q_i, v_i^{\text{temp}})$, and
$r_i^{\mathrm{s}} = \operatorname{rank}_{\mathcal{T}}(q_i, v_i^{\text{sem}})$.
Then
\begin{equation}
\text{Acc}_{\text{tri}}
= \frac{1}{N} \sum_{i=1}^{N} 
\mathbf{1}\bigl[
  r_i^{\mathrm{m}} < r_i^{\mathrm{t}}
  \land
  r_i^{\mathrm{m}} < r_i^{\mathrm{s}}
\bigr].
\end{equation}

\paragraph{Relative-order histogram.}
Beyond scalar accuracies, we also report the empirical distribution of
three-way orders within each triplet. For example:
\[
\begin{aligned}
&v^{\text{main}} \succ v^{\text{temp}} \succ v^{\text{sem}},\\
&v^{\text{temp}} \succ v^{\text{main}} \succ v^{\text{sem}},\\
&v^{\text{sem}} \succ v^{\text{main}} \succ v^{\text{temp}},\ \text{etc.}
\end{aligned}
\]
where $a \succ b$ means $\mathrm{rank}_{\mathcal{T}}(q,a) < \mathrm{rank}_{\mathcal{T}}(q,b)$.
This histogram reveals whether the model tends to systematically favor temporal or semantic hard negatives over the main video, and whether errors are dominated by temporal confusion (main ranked behind $v^{\text{temp}}$) or semantic confusion (main ranked behind $v^{\text{sem}}$).

{These triplet-level metrics are specific to the GenState-AI design and complement standard Recall@K: they directly measure whether a model correctly distinguishes the intended main state transition from carefully controlled temporal and semantic alternatives.}

\section{Experiments}

\subsection{Implementation Details}

We evaluate VCF-Lik, Qwen3-VL (2B/8B), and LanguageBind under a unified 5-frame full-video protocol. {Qwen3-VL uses scale-matched Qwen3-VL-Embedding-2B/8B first-stage retrievers with corresponding 2B/8B second-stage rerankers; LanguageBind uses the same 5 uniformly sampled frames for comparability with Qwen3-VL and the last-frame ablation.} {For full-pool Qwen3-VL retrieval, the embedding model ranks all videos in the evaluated subset and passes the top-$K$ candidates with $K=20$ to the corresponding reranker; the final ranking reranks these 20 candidates and retains the remaining candidates in first-stage order.} {For Recall@K, the mapped main clip is the sole positive, and the candidate pool contains all videos in the corresponding evaluated subset for all three baselines; all 195 queries are retained.} Acc$_{\text{temp}}$, Acc$_{\text{sem}}$, and Acc$_{\text{tri}}$ are computed over the query-specific triplet as defined in Sec.~\ref{sec:tripletMetrics}. {These metrics use the 154 complete triplets and exclude 41 incomplete groups.} All Avg.\ rows are macro averages over the four subsets.

\subsection{Overall Results on GenState-AI}

Table~\ref{tab:genstate_triplet_metrics} summarizes the performance of the four retrieval systems across the GenState-AI benchmark. {All Recall results use the corresponding complete subset pool for every baseline. Appendix~\ref{sec:appendix_global_pool} reports an optional 956-video VCF-Lik stress test, and Appendix~\ref{sec:appendix_languagebind_frames} reports LanguageBind frame-sampling sensitivity.}

\paragraph{Scaling and Remaining Challenges.}
The results show that GenState-AI is sensitive to model capability while still exposing structured failures.
{Scaling the Qwen3-VL configuration from 2B to 8B, with scale-matched embedding and reranking models, improves the macro-average R@1 from 0.539 to 0.581, while Acc$_{\text{tri}}$ increases only modestly from 0.693 to 0.721.}
This scale-matched comparison changes both the embedding and reranking components rather than isolating the reranker effect; higher full-pool retrieval scores do not automatically imply reliable triplet ordering.
{However, the improvement is not uniform across domains: on \textit{Manipulation}, Qwen3-VL (8B) obtains a lower R@1 than Qwen3-VL (2B) (0.304 vs. 0.348), while both achieve the same triplet consistency (0.524). In contrast, VCF-Lik performs substantially better, reaching 0.478 R@1 and 0.810 Acc$_{\text{tri}}$.}
Thus, the larger scale-matched Qwen3-VL configuration attains higher macro-average retrieval scores, while fine-grained action-dependent temporal grounding remains challenging.

\paragraph{Performance Comparison.}
Table~\ref{tab:genstate_triplet_metrics} shows that no single baseline dominates all evaluation axes.
{\textbf{Qwen3-VL (8B)} achieves the best macro-average retrieval performance (R@1: 0.581; R@5: 0.960; R@10: 0.985), suggesting that the larger scale-matched Qwen3-VL configuration is effective at selecting a single best candidate when the correct video survives first-stage retrieval.}
\textbf{LanguageBind}, however, obtains the best results on the \textit{Cartoon} subset (R@1: 0.833; Acc$_{\text{tri}}$: 0.833), indicating that its language-aligned video embedding space is robust for stylized content.
In contrast, \textbf{VCF-Lik} remains strongest on \textit{Manipulation} in both R@1 (0.478) and Acc$_{\text{tri}}$ ({0.810}), highlighting that likelihood-based matching can better preserve fine-grained action and end-state ordering when hand-object interactions are subtle and action-dependent.
Overall, the systems exhibit complementary strengths across domains, indicating that GenState-AI captures diverse failure modes beyond static appearance matching.

\paragraph{Triplet Consistency.}
Across subsets, semantic discrimination is generally easier than temporal discrimination for all baselines (i.e., Acc$_{\text{sem}}$ $>$ Acc$_{\text{temp}}$ in most cases), aligning with the intuition that substituting objects or attributes often changes global appearance more than subtle end-state differences.
This pattern is especially clear for \textbf{LanguageBind}: it achieves a high macro-average Acc$_{\text{sem}}$ (0.926), but its Acc$_{\text{temp}}$ remains much lower (0.685), with a pronounced gap on \textit{Manipulation} (0.571 vs.\ 0.857).
{\textbf{Qwen3-VL (8B)} narrows this gap on average (Acc$_{\text{temp}}$: 0.770; Acc$_{\text{sem}}$: 0.924), but still suffers on \textit{Manipulation}, where Acc$_{\text{tri}}$ is 0.524.}
These results reveal a distinction between broad semantic matching and decisive temporal endpoint verification: stronger models improve retrieval, yet paired temporal negatives still expose subset-specific weaknesses.
Overall, the results support the central motivation of GenState-AI: modern baselines can retrieve globally plausible videos, and the larger scale-matched Qwen3-VL configuration attains higher macro-average top-rank accuracy, but reliable state-aware grounding remains uneven across domains.

\paragraph{Value of Triplet Diagnostics.} Recall@K and triplet consistency capture complementary aspects of performance. LanguageBind attains macro-average R@5/R@10 of 0.916/0.958 on all 195 queries, whereas its Acc$_{\text{temp}}$ is 0.685 on the 154 complete triplets. These aggregate results concern different query sets, but illustrate that high global recall alone does not ensure correct temporal ordering. Despite lower macro-average Recall@K than Qwen3-VL, VCF-Lik achieves higher Manipulation Acc$_{\text{tri}}$ (0.810 vs. 0.524 for both Qwen3-VL variants), showing what standard retrieval metrics alone would miss about state-sensitive ordering.

\subsection{Diagnostic Ordering and Qualitative Trends}

Beyond scalar triplet consistency, LanguageBind's relative-order distribution shows that its top-ranked errors are dominated by temporal hard negatives. Among the 154 complete triplets, the main video ranks first in 107 (69.5\%) cases, a temporal negative in 45 (29.2\%), and a semantic negative in 2 (1.3\%). Temporal negatives rank first most often in Toy and Manipulation (38.1\% in each). Appendix~\ref{sec:appendix_diagnostics} provides the full ordering distributions and representative failures, illustrating the same challenge: a plausible clip can outrank the target despite conflicting with the queried motion or action.

\subsection{Ablation Study: Does the End-State Alone Suffice?}
\label{sec:ablation_last_frame}
Since our dataset emphasizes explicit final states, a natural question is whether models can solve it by inspecting only the terminal frame.
We therefore compare the default full-video setting (5 uniformly sampled frames) with a \textit{Last Frame Only} setting for Qwen3-VL and LanguageBind.
This ablation does not by itself prove temporal sequence reasoning; instead, it tests whether terminal-frame appearance matching is sufficient for the evaluated models.
Full subset-level results are provided in Appendix~\ref{sec:appendix_ablation}.

{At the macro-average level, full-video input improves R@1 and Acc$_{\text{tri}}$ for all evaluated systems, although subset-level metrics do not improve uniformly (Appendix~\ref{sec:appendix_ablation}). For Qwen3-VL (8B), R@1/Acc$_{\text{tri}}$ fall from 0.581/0.721 to 0.399/0.515 under last-frame-only input; LanguageBind falls from 0.559/0.673 to 0.435/0.574. On Toy, Qwen3-VL (8B) drops from 0.667 to 0.238 in R@1 and from 0.762 to 0.238 in Acc$_{\text{temp}}$. These results show that terminal-frame appearance alone is insufficient for this benchmark.}

\section{Conclusion}

{
We introduced GenState-AI, a prompt-controlled benchmark of AI-generated videos and verified state-change queries across four visual domains. Its evaluation separates standard retrieval over complete subset-level video pools from diagnostic ordering over query groups with complete main, temporal-negative, and semantic-negative mappings. This distinction reveals that a model may retrieve the correct video near the top globally while still preferring a paired clip with an incorrect state transition. The scale-matched Qwen3-VL (8B) configuration improves macro-average retrieval and triplet consistency over 2B, but temporal/state confusions remain. Qwen3-VL (8B) achieves the strongest overall retrieval and triplet consistency, whereas VCF-Lik performs notably better on Manipulation triplets, demonstrating complementary strengths across scoring paradigms. GenState-AI is therefore intended as a focused diagnostic benchmark rather than an exhaustive measure of video understanding. Its verified candidate pools, controlled hard negatives, and explicit evaluation protocol provide a reproducible basis for developing retrieval systems that better distinguish semantic relevance from state-transition correctness.
}

\clearpage
\section*{Limitations}
GenState-AI is prompt-controlled rather than pixel-controlled. Although hard-negative prompts are minimally edited and all clips are manually verified, independently generated videos may still vary in low-level appearance, lighting, pose, camera framing, or generation artifacts.
All clips use Wan2.2-TI2V-5B, so residual generator-specific cues may remain and cross-generator generalization remains untested.
The current benchmark focuses on four common state-centric subsets and 195 verified queries; it is intended as a diagnostic testbed rather than a comprehensive coverage of all event types, domains, or video generation models.
Finally, our experiments cover representative open-source retrieval paradigms, including likelihood scoring, embedding--reranking, and dual-encoder matching, but do not exhaustively evaluate all closed-source or larger-scale multimodal systems.

\section*{Acknowledgments}
This work was supported by the National Natural Science Foundation of China (No. 62376178).

\bibliography{main}

\appendix
\section{Qualitative Dataset Examples}
\label{sec:appendix_examples}

\begin{figure*}[!t]
\centering
\setlength{\fboxsep}{2pt}
\fbox{%
\begin{minipage}{0.47\linewidth}
  \centering
  \textbf{Home Group Example}
  \includegraphics[width=.98\linewidth]{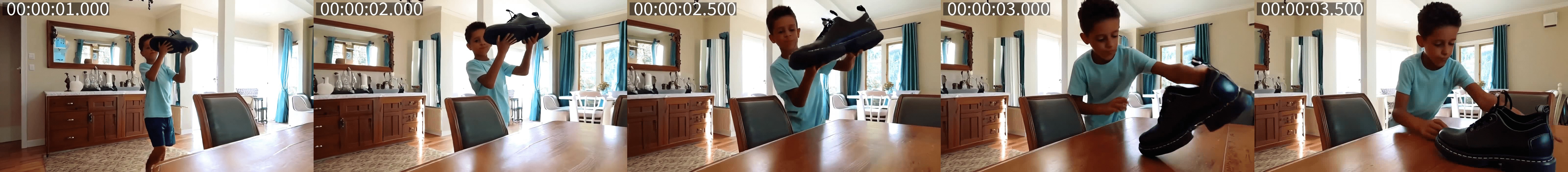} \\[-2pt]
  \scriptsize Main
  \includegraphics[width=.98\linewidth]{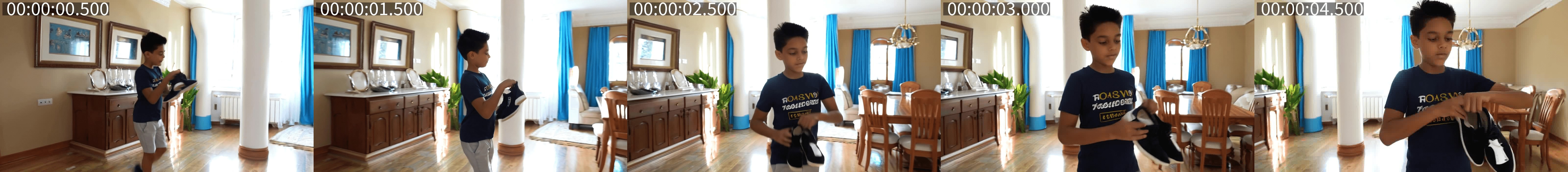} \\[-2pt]
  \scriptsize Temporal Negative
  \includegraphics[width=.98\linewidth]{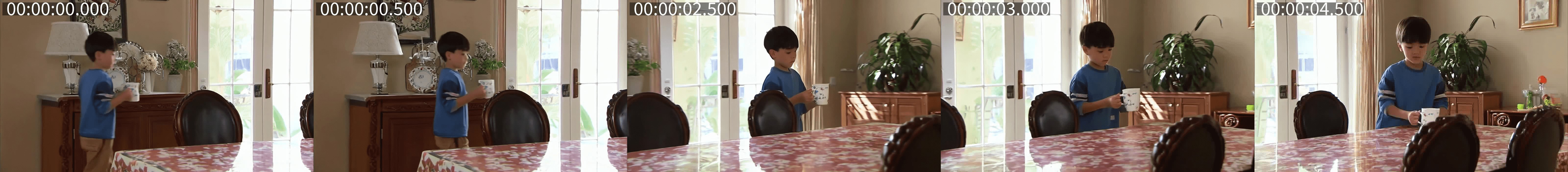} \\[-2pt]
  \scriptsize Semantic Negative
\end{minipage}%
}
\hfill
\fbox{%
\begin{minipage}{0.47\linewidth}
  \centering
  \textbf{Toy Group Example}
  \includegraphics[width=.98\linewidth]{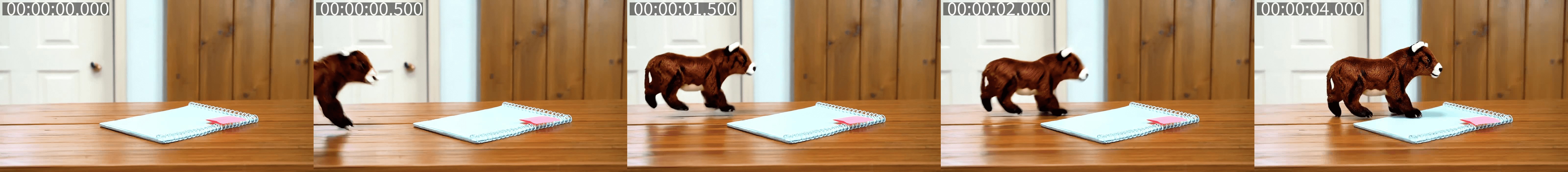} \\[-2pt]
  \scriptsize Main
  \includegraphics[width=.98\linewidth]{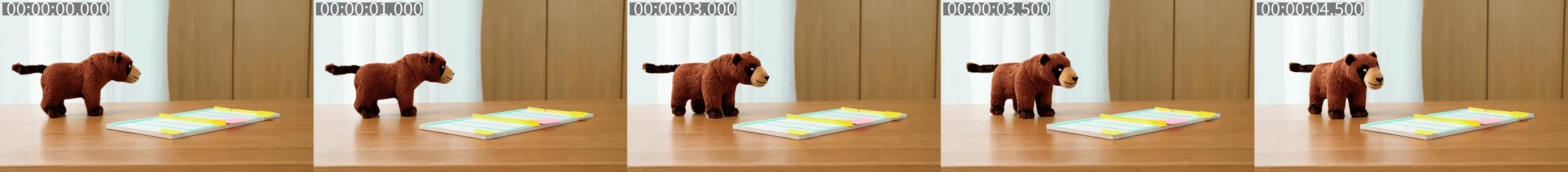} \\[-2pt]
  \scriptsize Temporal Negative
  \includegraphics[width=.98\linewidth]{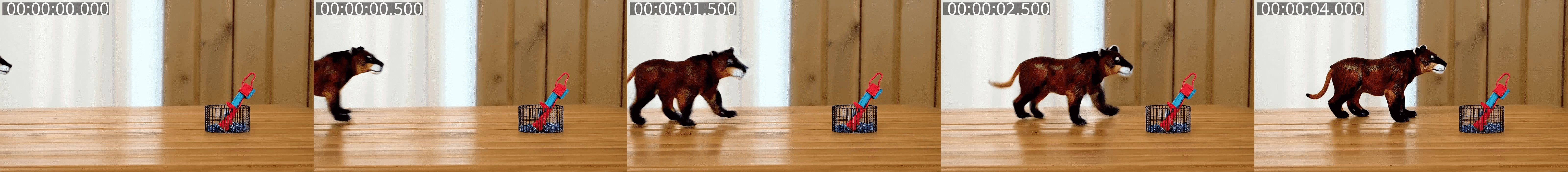} \\[-2pt]
  \scriptsize Semantic Negative
\end{minipage}%
}

\vspace{3pt}

\fbox{%
\begin{minipage}{0.47\linewidth}
  \centering
  \textbf{Cartoon Group Example}
  \includegraphics[width=.98\linewidth]{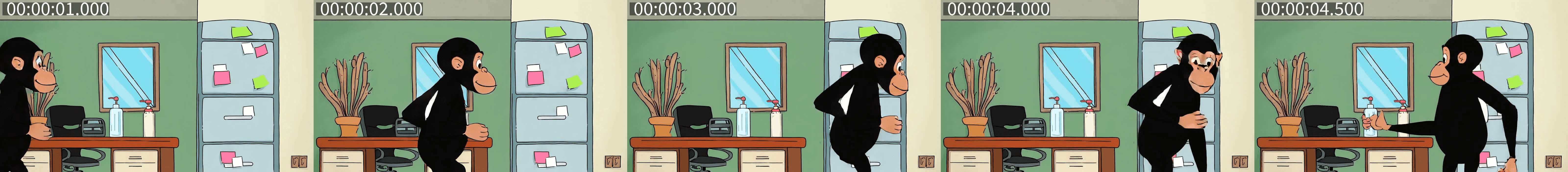} \\[-2pt]
  \scriptsize Main
  \includegraphics[width=.98\linewidth]{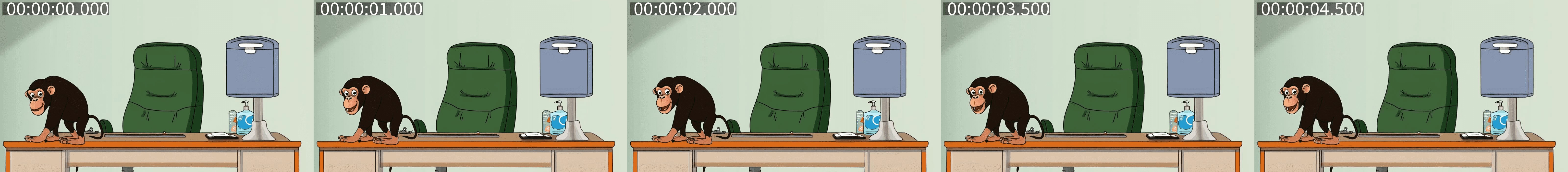} \\[-2pt]
  \scriptsize Temporal Negative
  \includegraphics[width=.98\linewidth]{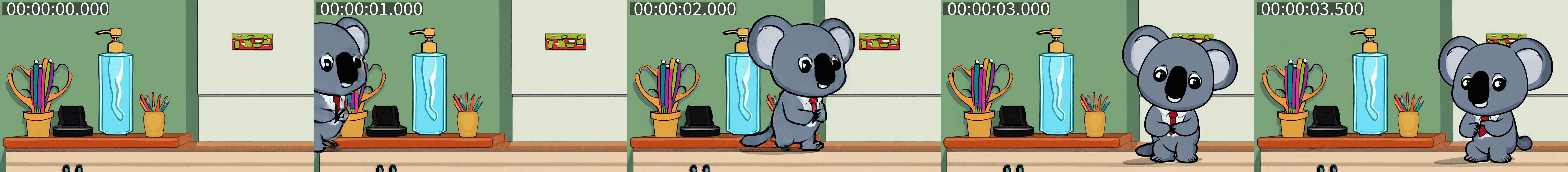} \\[-2pt]
  \scriptsize Semantic Negative
\end{minipage}%
}
\hfill
\fbox{%
\begin{minipage}{0.47\linewidth}
  \centering
  \textbf{Manipulation Group Example}
  \includegraphics[width=.98\linewidth]{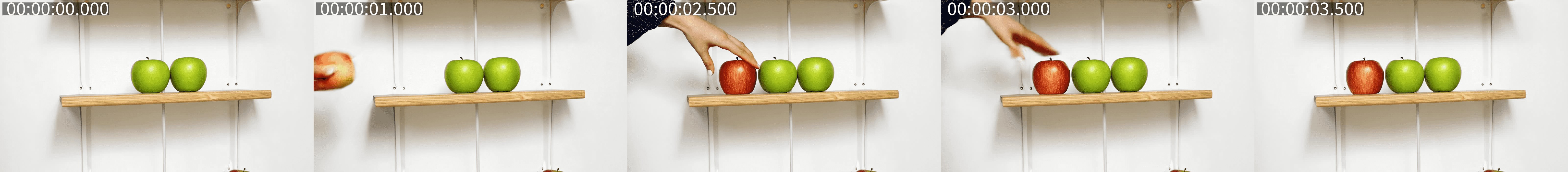} \\[-2pt]
  \scriptsize Main
  \includegraphics[width=.98\linewidth]{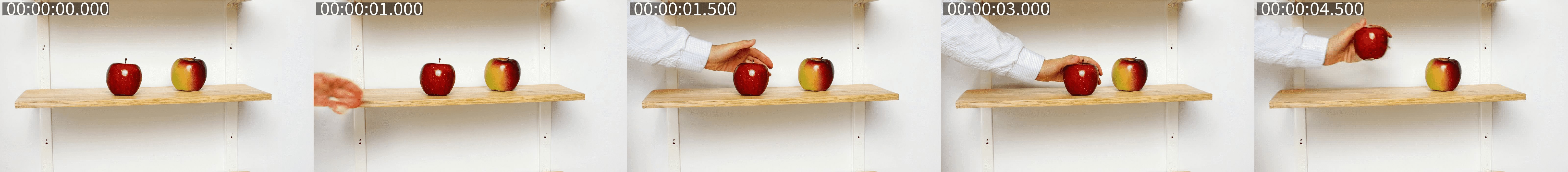} \\[-2pt]
  \scriptsize Temporal Negative
  \includegraphics[width=.98\linewidth]{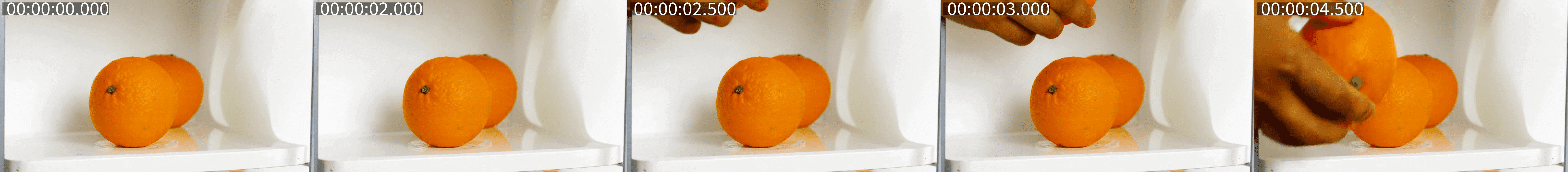} \\[-2pt]
  \scriptsize Semantic Negative
\end{minipage}%
}
\caption{Qualitative examples of generated videos and paired hard negatives across all four subsets. Each image is a horizontally concatenated video strip; each group shows the main video, temporal hard negative, and semantic hard negative.}
\label{fig:benchmark_viz_app}
\end{figure*}

Figure~\ref{fig:benchmark_viz_app} provides one representative triplet from each subset. These examples are intended to make the construction protocol visually inspectable: the temporal negative targets a conflicting state-transition outcome, while the semantic negative changes the queried object, attribute, or action target.

Because the clips are generated independently from edited prompts, they should be read as prompt-controlled rather than pixel-controlled counterfactuals. The examples show that the paired videos preserve the broad scene and task setting while still exhibiting natural generation-level variation in pose, lighting, object appearance, and composition.

\section{Full Last-Frame Ablation Results}
\label{sec:appendix_ablation}

\begin{table*}[!t]
\centering
\caption{Full subset-level ablation comparing full-video context (5 frames) against static last-frame context. {Recall@K uses all videos in each corresponding subset pool and retains all 195 queries, whereas the triplet accuracies use the 154 complete query-specific triplets; 41 incomplete query groups are excluded.}}
\label{tab:last_frame_ablation_full}
\footnotesize
\setlength{\tabcolsep}{4pt}
\begin{tabular}{llccccccc}
\toprule
Subset & Model & Context & R@1 & R@5 & R@10 & Acc$_{temp}$ & Acc$_{sem}$ & Acc$_{tri}$ \\
\midrule
\multirow{6}{*}{Cartoon}
& Qwen3-VL (2B) & Full & {0.708} & {1.000} & {1.000} & {0.792} & {0.875} & {0.750} \\
& Qwen3-VL (2B) & Last & {0.625} & {0.833} & {0.917} & {0.708} & {0.917} & {0.667} \\
& Qwen3-VL (8B) & Full & {0.667} & {1.000} & {1.000} & {0.792} & {0.958} & {0.750} \\
& Qwen3-VL (8B) & Last & {0.625} & {0.958} & {1.000} & {0.667} & {0.958} & {0.667} \\
& LanguageBind & Full & 0.833 & 0.958 & 1.000 & 0.833 & 1.000 & 0.833 \\
& LanguageBind & Last & 0.750 & 0.958 & 0.958 & 0.750 & 0.917 & 0.750 \\
\midrule
\multirow{6}{*}{Home}
& Qwen3-VL (2B) & Full & {0.528} & {0.850} & {0.953} & {0.852} & {0.943} & {0.830} \\
& Qwen3-VL (2B) & Last & {0.409} & {0.780} & {0.921} & {0.739} & {0.898} & {0.705} \\
& Qwen3-VL (8B) & Full & {0.685} & {0.929} & {0.984} & {0.909} & {0.977} & {0.898} \\
& Qwen3-VL (8B) & Last & {0.472} & {0.835} & {0.953} & {0.761} & {0.898} & {0.727} \\
& LanguageBind & Full & 0.441 & 0.890 & 0.969 & 0.716 & 0.989 & 0.716 \\
& LanguageBind & Last & 0.299 & 0.717 & 0.858 & 0.614 & 0.830 & 0.545 \\
\midrule
\multirow{6}{*}{Manip.}
& Qwen3-VL (2B) & Full & {0.348} & {0.783} & {0.957} & {0.524} & {0.762} & {0.524} \\
& Qwen3-VL (2B) & Last & {0.174} & {0.609} & {0.957} & {0.524} & {0.857} & {0.476} \\
& Qwen3-VL (8B) & Full & {0.304} & {0.957} & {0.957} & {0.619} & {0.857} & {0.524} \\
& Qwen3-VL (8B) & Last & {0.261} & {0.696} & {0.870} & {0.476} & {0.762} & {0.429} \\
& LanguageBind & Full & 0.391 & 0.913 & 0.913 & 0.571 & 0.857 & 0.524 \\
& LanguageBind & Last & 0.261 & 0.826 & 0.957 & 0.571 & 0.905 & 0.476 \\
\midrule
\multirow{6}{*}{Toy}
& Qwen3-VL (2B) & Full & {0.571} & {0.952} & {0.952} & {0.667} & {0.905} & {0.667} \\
& Qwen3-VL (2B) & Last & {0.476} & {0.952} & {0.952} & {0.524} & {0.810} & {0.524} \\
& Qwen3-VL (8B) & Full & {0.667} & {0.952} & {1.000} & {0.762} & {0.905} & {0.714} \\
& Qwen3-VL (8B) & Last & {0.238} & {0.857} & {0.905} & {0.238} & {0.810} & {0.238} \\
& LanguageBind & Full & 0.571 & 0.905 & 0.952 & 0.619 & 0.857 & 0.619 \\
& LanguageBind & Last & 0.429 & 0.810 & 0.857 & 0.524 & 0.810 & 0.524 \\
\midrule
\multirow{6}{*}{Avg.}
& Qwen3-VL (2B) & Full & {0.539} & {0.896} & {0.965} & {0.709} & {0.871} & {0.693} \\
& Qwen3-VL (2B) & Last & {0.421} & {0.793} & {0.937} & {0.624} & {0.870} & {0.593} \\
& Qwen3-VL (8B) & Full & {0.581} & {0.960} & {0.985} & {0.770} & {0.924} & {0.721} \\
& Qwen3-VL (8B) & Last & {0.399} & {0.836} & {0.932} & {0.536} & {0.857} & {0.515} \\
& LanguageBind & Full & 0.559 & 0.916 & 0.958 & 0.685 & 0.926 & 0.673 \\
& LanguageBind & Last & 0.435 & 0.828 & 0.908 & 0.615 & 0.865 & 0.574 \\
\bottomrule
\end{tabular}
\end{table*}

Table~\ref{tab:last_frame_ablation_full} reports the complete subset-level results for the last-frame ablation. All Avg.\ rows are macro averages over the four subsets. {Recall@K uses each corresponding complete subset pool and all 195 query descriptions, while the triplet accuracies use the 154 complete query-specific triplets.}

The full table expands the summary reported in the main experiments. Across models, replacing the 5-frame video input with only the final frame generally reduces both retrieval recall and triplet consistency, especially on state-centric subsets such as Toy and Home. This supports the conservative conclusion that terminal-frame appearance alone is insufficient for reliably solving GenState-AI, although it does not by itself prove that models perform full temporal-order reasoning.

\section{LanguageBind Frame-Sampling Sensitivity}
\label{sec:appendix_languagebind_frames}

Table~\ref{tab:languagebind_frame_sensitivity} compares LanguageBind with the unified 5-frame protocol used in the main experiments against its 8-frame default setting. We treat this as a frame-sampling sensitivity analysis rather than a separate main baseline.

\begin{table*}[htbp]
\centering
\caption{LanguageBind frame-sampling sensitivity. The 5-frame setting is used in the main tables for protocol consistency. {Recall@K uses each corresponding complete subset pool; triplet accuracies use the 154 complete triplets.}}
\label{tab:languagebind_frame_sensitivity}
\scriptsize
\setlength{\tabcolsep}{2pt}
\resizebox{1.1\columnwidth}{!}{%
\begin{tabular}{llcccccc}
\toprule
Subset & Frames & R@1 & R@5 & R@10 & Acc$_{temp}$ & Acc$_{sem}$ & Acc$_{tri}$ \\
\midrule
Toy & 5 & 0.571 & 0.905 & 0.952 & 0.619 & 0.857 & 0.619 \\
Toy & 8 & 0.571 & 0.905 & 0.952 & 0.619 & 0.857 & 0.619 \\
Cartoon & 5 & 0.833 & 0.958 & 1.000 & 0.833 & 1.000 & 0.833 \\
Cartoon & 8 & 0.750 & 1.000 & 1.000 & 0.792 & 1.000 & 0.792 \\
Home & 5 & 0.441 & 0.890 & 0.969 & 0.716 & 0.989 & 0.716 \\
Home & 8 & 0.504 & 0.921 & 0.984 & 0.750 & 0.989 & 0.750 \\
Manip. & 5 & 0.391 & 0.913 & 0.913 & 0.571 & 0.857 & 0.524 \\
Manip. & 8 & 0.304 & 0.870 & 0.957 & 0.476 & 0.905 & 0.429 \\
\midrule
Avg. & 5 & 0.559 & 0.916 & 0.958 & 0.685 & 0.926 & 0.673 \\
Avg. & 8 & 0.532 & 0.924 & 0.973 & 0.659 & 0.938 & 0.647 \\
\bottomrule
\end{tabular}
}
\end{table*}

The 5-frame setting is used in the main tables so that LanguageBind is evaluated under the same visual evidence budget as VCF-Lik and Qwen3-VL. The 8-frame setting is retained here because it reflects the model's default usage and helps estimate whether the main conclusion depends on this implementation choice. The difference is modest at the macro-average level, but subset-level changes are visible: 8 frames improve Home while reducing Cartoon and Manipulation, suggesting that frame sampling can interact with scene type and generated-video style.

\section{{VCF-Lik Dataset-Wide Retrieval Stress Test}}
\label{sec:appendix_global_pool}

{Table~\ref{tab:vcf_global_pool} reports an optional 956-video cross-subset stress test that is not used in Table~\ref{tab:genstate_triplet_metrics}. Only Cartoon decreases because non-Cartoon candidates displace correct clips across the Recall cutoffs; Toy, Home, and Manipulation are unchanged.}
The subset-pool columns rank each query against videos in its corresponding subset, whereas the 956-video columns use the full dataset as the candidate pool.

\begin{table}[htbp]
\centering
\caption{VCF-Lik Recall using the Table~\ref{tab:genstate_triplet_metrics} subset pools versus all 956 videos.}
\label{tab:vcf_global_pool}
\scriptsize
\setlength{\tabcolsep}{2.5pt}
\renewcommand{\arraystretch}{0.92}
\resizebox{0.9\columnwidth}{!}{%
\begin{tabular}{lcccccc}
\toprule
& \multicolumn{3}{c}{Subset pool} & \multicolumn{3}{c}{956-video pool} \\
\cmidrule(lr){2-4}\cmidrule(lr){5-7}
Subset & R@1 & R@5 & R@10 & R@1 & R@5 & R@10 \\
\midrule
Toy & 0.381 & 0.714 & 0.857 & 0.381 & 0.714 & 0.857 \\
Cartoon & 0.250 & 0.708 & 0.708 & 0.208 & 0.417 & 0.625 \\
Home & 0.386 & 0.512 & 0.591 & 0.386 & 0.512 & 0.591 \\
Manip. & 0.478 & 0.870 & 0.957 & 0.478 & 0.870 & 0.957 \\
\bottomrule
\end{tabular}
}
\end{table}

\section{Diagnostic Ordering and Failure Visualizations}
\label{sec:appendix_diagnostics}

\begin{figure*}[!b]
    \centering
    \includegraphics[width=\textwidth]{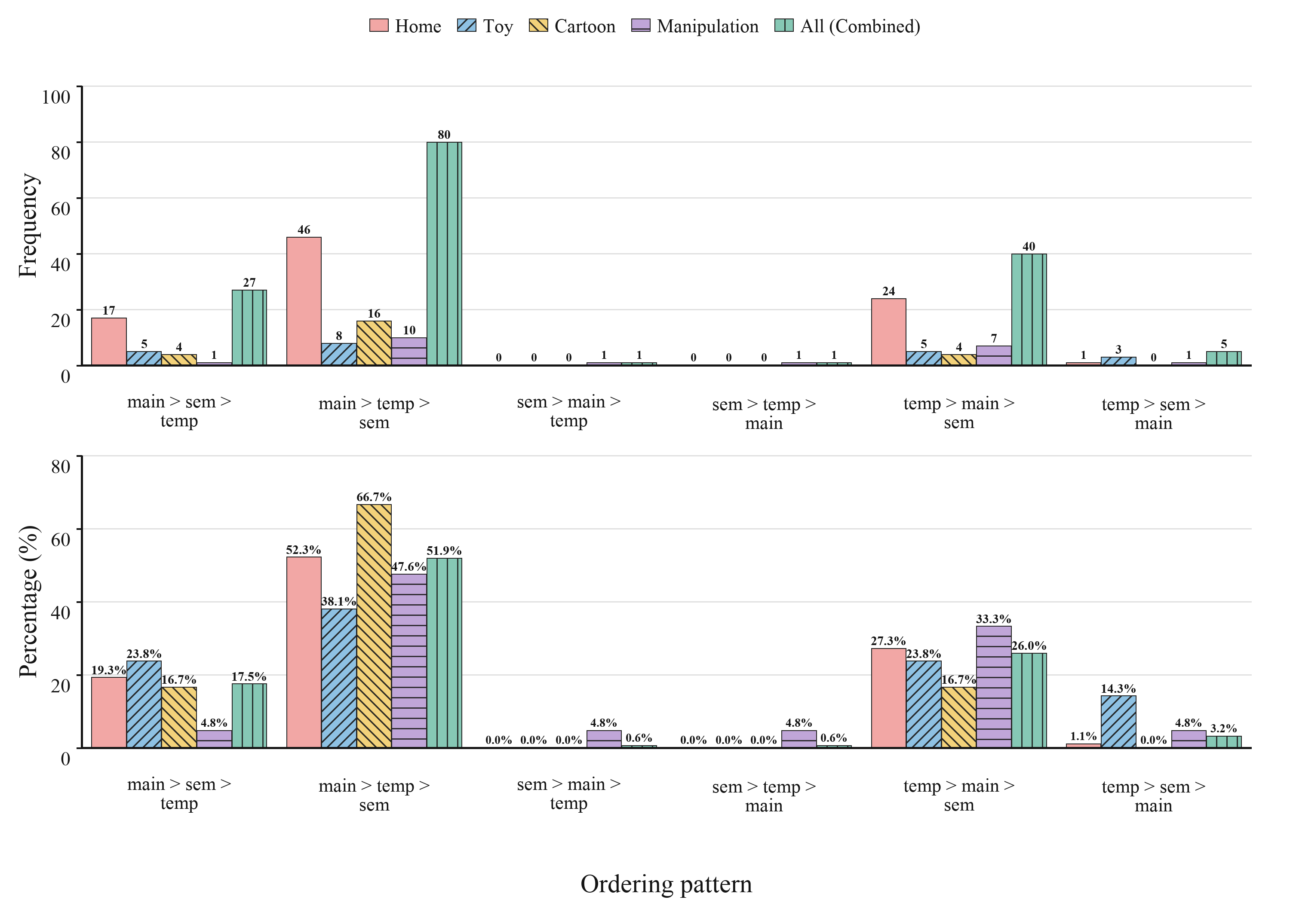}
    \caption{Comparison of relative-order distributions across subsets under {\textbf{LanguageBind}}. {Counts and percentages are computed from the complete triplets available for each subset (154 in total).}}
    \label{fig:ordering_subset}
\end{figure*}

{Figure~\ref{fig:ordering_subset} summarizes the relative-order distributions of the 154 complete triplets under LanguageBind. The desired ordering places the main video ahead of both negatives; deviations reveal whether the model is more often confused by temporal/state conflicts or semantic distractors. Across all triplets, the main video is ranked first in 107/154 (69.5\%) cases, a temporal negative in 45/154 (29.2\%), and a semantic negative in 2/154 (1.3\%).}

\begin{figure*}[!t]
\centering
\setlength{\fboxsep}{2pt}
\fbox{%
\begin{minipage}{0.99\linewidth}
\centering
\textbf{Toy Failure Case}\par
\includegraphics[width=.98\linewidth]{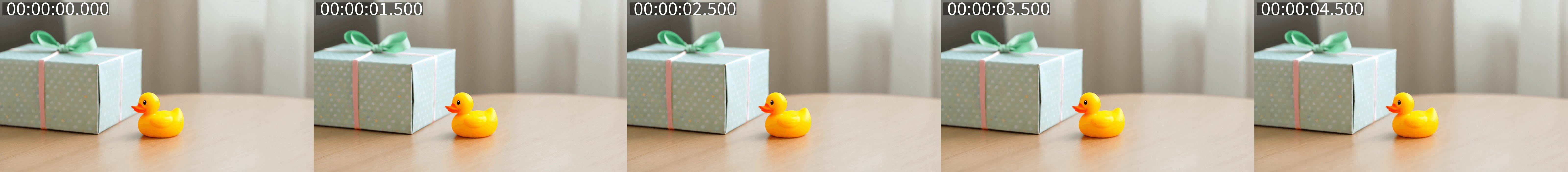}\\[-2pt]
{\scriptsize Rank 1: Temporal Negative}\par
\includegraphics[width=.98\linewidth]{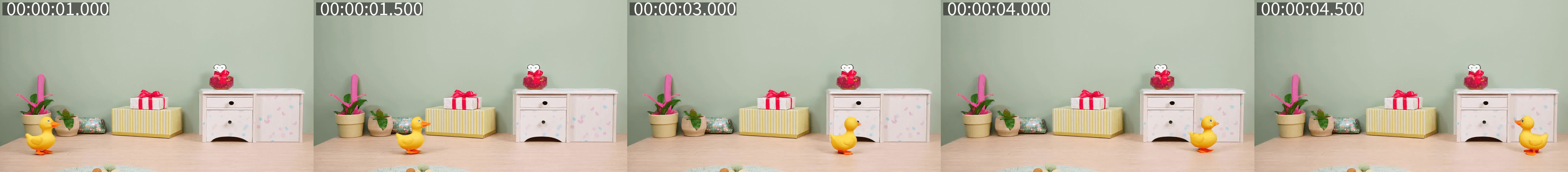}\\[-2pt]
{\scriptsize Rank 2 (Main): duck toy moves}\par
\includegraphics[width=.98\linewidth]{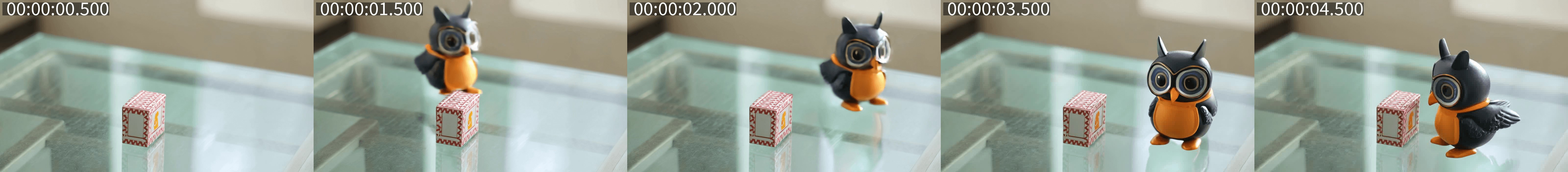}\\[-2pt]
{\scriptsize Rank 3: Semantic Negative}
\end{minipage}%
}%
\vspace{3pt}
\fbox{%
\begin{minipage}{0.99\linewidth}
\centering
\textbf{Home Failure Case}\par
\includegraphics[width=.98\linewidth]{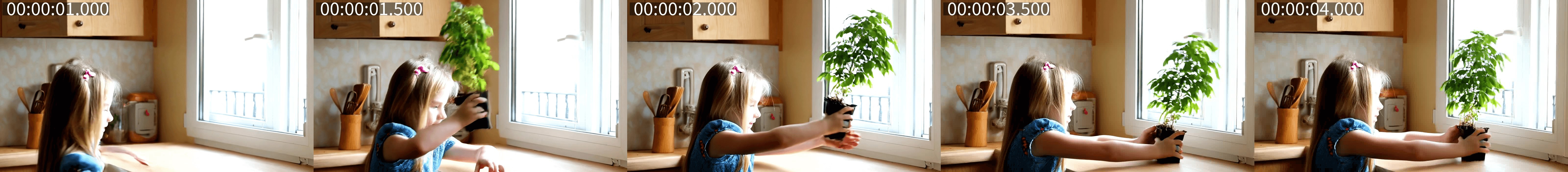}\\[-2pt]
{\scriptsize Rank 1: Semantic Negative}\par
\includegraphics[width=.98\linewidth]{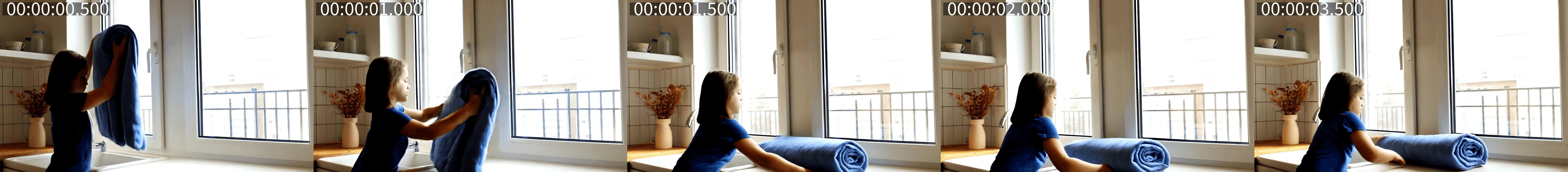}\\[-2pt]
{\scriptsize Rank 2: Temporal Negative}\par
\includegraphics[width=.98\linewidth]{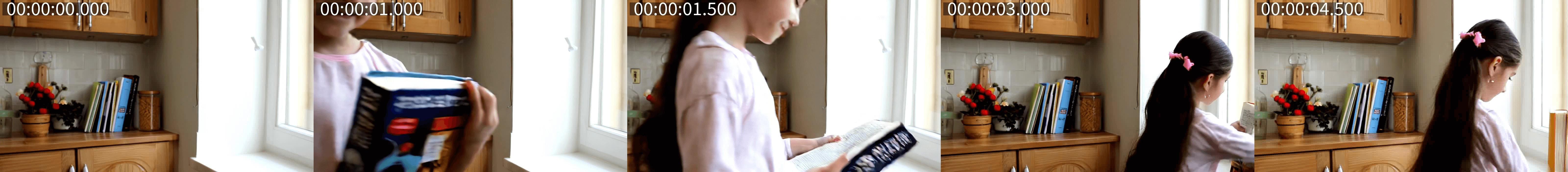}\\[-2pt]
{\scriptsize Rank 3 (Main): picking up the book}
\end{minipage}%
}%
\caption{{Representative triplet-ordering failures under \textbf{VCF-Lik}; higher-ranked negatives appear above the main video. In \textbf{Toy}, Rank~1 is a temporal negative showing the same rubber duck moving from left to right beside a gift box, while the Rank~2 target shows the duck moving from right to left; Rank~3 is a semantic negative featuring a jumping cartoon owl. Thus, the same-object clip with opposite motion outranks the target. In \textbf{Home}, Rank~1 is a semantic negative showing a girl moving a potted plant to a windowsill, Rank~2 is a temporal negative showing her lifting and laying down a blue blanket, and the Rank~3 target shows her picking up a book. This ordering illustrates difficulty distinguishing the target action from hard negatives in a similar visual context.}}
\label{fig:failure_cases}
\end{figure*}

Under LanguageBind, temporal negatives rank first in 38.1\% of complete triplets in both Toy and Manipulation. Figure~\ref{fig:failure_cases} shows two VCF-Lik errors: a same-object, opposite-motion Toy negative outranks the target, as does an action-mismatched Home negative in a similar scene.

\end{document}